\documentclass[runningheads]{llncs}

\usepackage{amsmath}
\usepackage{amssymb}
\usepackage{xcolor}
\usepackage{algorithm}
\usepackage{algorithmic}
\usepackage{subcaption}
\usepackage{graphicx}
\usepackage[normalem]{ulem}
\usepackage{cleveref}

\newcommand{\R}{\mathbb{R}}

\newcommand{\cI}{\mathcal{I}}

\newcommand{\lc}{\textrm{left}}
\newcommand{\rc}{\textrm{right}}
\newcommand{\ch}{\textrm{child}}

\newcommand{\Impurity}{\textrm{I}}
\newcommand{\decrease}{\Delta_{\cI}}
\newcommand{\node}{v} 
\newcommand{\vertex}{u} 
\newcommand{\tree}{t}
\newcommand{\forest}{F}
\newcommand{\depth}{P}
\newcommand{\distance}{D}
\newcommand{\added}{\mathcal{A}}
\newcommand{\toadd}{\mathcal{B}}
\newcommand{\size}{N}
\newcommand{\fs}{X}

\definecolor{martincolor}{rgb}{1.0, 0.0, 0.0}

\definecolor{bastiancolor}{rgb}{0.2, 0.6, 0.9}

\definecolor{christelcolor}{rgb}{0.1, 0.8, 0.1}

\begin{document}
%


\title{Feature graphs for interpretable unsupervised tree ensembles: centrality, interaction, and application in disease subtyping
}

\titlerunning{Feature graphs for interpretable unsupervised tree ensembles}
%
\author{Christel Sirocchi\inst{1}\orcidID{0000-0002-5011-3068} \and
Martin Urschler \inst{2}\orcidID{0000-0001-5792-3971} \and
Bastian Pfeifer\inst{2}\orcidID{0000-0001-7035-9535} }
\authorrunning{C. Sirocchi et al.}
%
\institute{Department of Pure and Applied Sciences \\ University of Urbino, Italy \and
Institute for Medical Informatics, Statistics and Documentation \\ Medical University of Graz, Austria}
\maketitle              
\begin{abstract}
Interpretable machine learning has emerged as central in leveraging artificial intelligence within high-stakes domains such as healthcare, where understanding the rationale behind model predictions is as critical as achieving high predictive accuracy. In this context, feature selection assumes a pivotal role in enhancing model interpretability by identifying the most important input features in black-box models. While random forests are frequently used in biomedicine for their remarkable performance on tabular datasets, the accuracy gained from aggregating decision trees comes at the expense of interpretability.
Consequently, feature selection for enhancing interpretability in random forests has been extensively explored in supervised settings. However, its investigation in the unsupervised regime remains notably limited.
To address this gap, the study introduces novel methods to construct feature graphs from unsupervised random forests and feature selection strategies to derive effective feature combinations from these graphs.
Feature graphs are constructed for the entire dataset as well as individual clusters leveraging the parent-child node splits within the trees, such that feature centrality captures their relevance to the clustering task, while edge weights reflect the discriminating power of feature pairs.
Graph-based feature selection methods are extensively evaluated on synthetic and benchmark datasets both in terms of their ability to reduce dimensionality while improving clustering performance, as well as to enhance model interpretability. 
An application on omics data for disease subtyping identifies the top features for each cluster, showcasing the potential of the proposed approach to enhance interpretability in clustering analyses and its utility in a real-world biomedical application.
 

\end{abstract}

\section{Introduction}
Interpretable machine learning has become a predominant concern across diverse domains since understanding the reasoning behind model predictions is widely considered at least as important as achieving high predictive accuracy~\cite{reddy2022explainability}.
In this context, decision trees stand out as interpretable models that offer transparency in the decision-making process.
However, their interpretability diminishes as they aggregate into tree ensembles. 
Random forests exhibit remarkable performance, particularly in tabular data, often outperforming deep learning techniques~\cite{grinsztajn2022tree}. This advantage is particularly notable in domains like biomedicine, where datasets are commonly organised in tabular form. 
Hence, the challenge lies in balancing the interpretability of individual decision trees with the enhanced predictive power achieved through ensemble methods.

As machine learning techniques advance, so does the field of eXplainable AI (XAI) to offer methods for interpretable machine learning, aimed at uncovering the underlying explanatory factors of a black-box model's prediction. XAI has introduced various explanation techniques, ranging from interpretable models that approximate input-output relations to methods that highlight the most influential input features in black-box predictions~\cite{bach2015pixel,ali2023enlightening}. In this context, feature selection efforts stand out as pivotal for improving model interpretability~\cite{bolon2015recent}. 
Feature selection for enhancing interpretability in random forests has been extensively explored in supervised settings~\cite{nembrini2018revival}, yet its investigation in the unsupervised regime remains limited.

Unsupervised random forests are highly effective in computing affinity matrices, which can be further analysed through clustering techniques, offering great potential for applications within the biomedical domain, particularly in disease subtyping.
Disease subtyping employs advanced clustering algorithms to stratify patients based on shared characteristics, such as clinical features and molecular profiles~\cite{leng2022benchmark}. This approach enables researchers to identify distinct subgroups within a disease, thereby enhancing the understanding of its underlying molecular complexity~\cite{rappoport2018multi,pfeifer2023parea}, ultimately facilitating personalised medical treatments. In this context, enhancing the largely unexplored interpretability of unsupervised random forests emerges as a critical pursuit, and is the focus of our work.


In this study, we improve the interpretability of unsupervised random forests by presenting a novel method for constructing feature graphs via leveraging the parent-child node splits within the trees. 
The feature graphs are constructed such that the centrality of features within the graph captures their relevance, and edges between features reflect the effectiveness of the feature pair in discriminating clusters. 
Feature graphs are constructed for the entire dataset or specific to each cluster, enabling the assessment of feature contributions within each cluster. Additionally, two feature selection strategies are introduced, a brute-force method and a greedy approach, aimed at selecting the top $k$ features in the dataset. 
The study extensively evaluates the effectiveness of the proposed graph-building and graph-mining methods on both synthetic and benchmark datasets. Feature selection is evaluated on multiple fronts, both in terms of its ability to reduce dimensionality and improve performance by filtering features for model training, and also to enhance model interpretability by identifying the most relevant features for each cluster and overall. An application on omics data for disease subtyping identifies the top features for each cluster, demonstrating the potential to enhance interpretability in clustering analyses and highlighting its utility in a real-world biomedical application. 

%

The article is structured as follows. Section~\ref{sec:PW} provides a comprehensive review of previous work on strategies for feature selection and model interpretability in unsupervised settings. Additionally, it outlines previous efforts in leveraging existing feature graphs to enhance random forests or constructing feature graphs from tree-based models. In Section~\ref{sec:MM}, the proposed methodology is introduced, beginning with a concise overview of unsupervised random forests and then detailing the method for constructing and mining feature graphs to estimate feature importance and perform feature selection. Section~\ref{sec:EV} describes the synthetic datasets generated within the study and the benchmark datasets utilised, as well as the experiments conducted to evaluate the proposed methods.
Section~\ref{sec:RD} presents and discusses the results of the investigation, showcasing a practical application of the method in the context of disease subtyping, and, finally, Section~\ref{sec:CF} concludes by summarising insights derived from the findings and outlining promising avenues for future research. 

\section{Previous Work}\label{sec:PW}

\subsection{Feature importance in unsupervised learning}

Unsupervised machine learning plays a predominant role in biomedical research, where clustering assists in grouping patients based on clinical or molecular features, supporting tasks such as disease subtyping or drug discovery for personalised medicine~\cite{ciriello2013emerging}. In disease subtyping, for instance, detected clusters typically consist of patients with different phenotypic characteristics, thereby requiring customised treatment. However, most of the existing disease subtyping methods do not infer the key biomarker characteristics for each cluster. A precise detection and selection of these features could enable clinicians to classify patients based on a small set of markers.


The field of XAI has contributed a plethora of explanation techniques to enhance model interpretability, for instance by identifying the most important input features in black-box predictions. However, research efforts aimed at improving interpretability in unsupervised models have been relatively limited, with a predominant focus on anomaly detection rather than clustering~\cite{kauffmann2022clustering}. 
Most XAI efforts have resorted to recasting the unsupervised problem into a supervised one. For instance, a common approach involves converting the unsupervised model into a functionally equivalent supervised model able to replicate the cluster assignments of the original clustering model~\cite{kauffmann2022clustering,montavon2020explaining}, so that available supervised XAI methods can be applied. 
However adopting a supervised perspective for an unsupervised problem risks compromising the inherent characteristics of unsupervised learning, such as uncovering hidden patterns or structures in data, particularly if the introduced labels do not accurately represent the underlying data structure.
Therefore, the demand for interpretability in unsupervised models remains high, as the main purpose of applying these models is often to better understand the underlying data~\cite{montavon2020explaining}. 
Identifying the features that most contributed to a clustering assignment stands out as one of the most valuable explanations to ultimately provide actionable insights. In the context of disease subtyping, relevant features can help identify signature genes and potential biomarkers, enabling the development of personalised screening tests~\cite{crippa2023characterization,sarkar2021machine}.


In unsupervised learning, methods for identifying relevant features have mainly been developed in the context of feature selection. These methods are often categorised based on their interaction with the clustering algorithm~\cite{dy2004feature}. Filter methods for feature selection assess the relevance of each feature using statistical metrics such as correlation or variance, independently of the target variable. Features are then sorted or thresholded, and a subset is chosen for further analysis or model training.
Filter methods are computationally efficient and scalable as they assess correlation and dependence among features without involving the model. However, they do not offer insights into how the model utilises features for clustering. In contrast, wrapper methods search for a feature subset such that the clustering algorithm, when trained on this subset, optimises a predefined criterion, thereby shedding light on the importance of features within the adopted clustering scheme~\cite{solorio2020review}. However, wrapper methods pose several challenges, such as the requirement for an appropriate criterion to guide the search and high computational costs due to the need to retrain the model on numerous candidate feature subsets~\cite{elghazel2015unsupervised}. 
Additionally, these methods may overlook the context of feature interactions, potentially identifying confounded features that lack direct relevance to the task at hand. The selection of features that, while yielding good performance, do not reflect any meaningful relationship between features and do not support any mechanistic explanation for the problem under study, greatly reduces model interpretability. 

To address this issue and to provide meaningful insights into feature importance, several strategies have emerged within the domain of tree ensembles. Some of these methods leverage existing graphs, while others compute feature graphs based on the data at hand. However, to the best of the authors' knowledge, all of these methods are adopted in a supervised setting.

\subsection{Leveraging feature graphs with random forests}
In the past decade, several studies, particularly within the realm of biomedicine, have leveraged networks representing known relationships among features to improve the quality of random forest models. These efforts aimed to address various challenges such as reducing over-fitting, and enhancing model robustness, data efficiency, interpretability, and coherence with prior knowledge. Most importantly, they sought to identify feature modules with greater biomedical relevance able to explain underlying physio-pathological mechanisms. These methodologies refine random forests by adjusting feature bagging (sub-sampling of features used for training each tree) and split feature selection (sub-sampling of features at each node during tree growth), or both. Such approaches proved effective, although they have been exclusively applied in a supervised setting. 

First, Dutkowski et al. developed the Network-Guided Forest method, where feature bagging is guided by graph search on a given biological network and, at each node, the split feature is chosen from the neighbours of the parent node~\cite{dutkowski2011protein}. 
Pan et al.~\cite{pan2013supervising} introduced the Mutual Information Network-guided random forests, which also select split features among the parent node's neighbours in the network, with probabilities proportional to corresponding edge weights. 
Andel et al.~\cite{andvel2015network} proposed the Network-Constrained Forest (NCF), which begins by sampling a feature, potentially relevant to the studied phenomenon, as a seed and selects the remaining features from a probabilistic distribution over the network, favouring those closer to the seed. 
Petralia et al.~\cite{petralia2015integrative} introduced iRafNet, an integrative random forest framework for gene regulatory network inference capable of incorporating information from diverse data sources. iRafNet derives weights from additional data sources such as protein-protein interactions, representing the prior belief of regulatory relationships for a given gene. Subsequently, using expression data as the primary source, random forests are trained to identify genes regulating the given gene, selecting features based on previously calculated weights.
Wang et al.~\cite{wang2018integration} presented Reweighted Random Survival Forests (RRSF) for survival prediction, which integrates the topological importance of genes assessed using directed random walk by selecting genes for node splitting based on their topological weight.
Larsen et al.~\cite{larsen2020novo} introduced Grand Forest, where feature bagging involves selecting a node uniformly at random and growing a subgraph using breadth-first search. Split variable selection ensures that each split variable is chosen only from variables connected to the parent node.
Pfeifer et al.~\cite{pfeifer2022multi} proposed Greedy Decision Forest subnetwork detection, which selects disease network modules based on multi-modal node features. The algorithm samples node features through random walks, generating decision trees from the visited nodes.
Finally, Tian et al.~\cite{tian2023graph} introduced Graph Random Forest (GRF), where the root node of each decision tree is determined through a data-driven approach, and features in the neighbourhood of any number of hops to the root node are considered to train each tree.


\subsection{Building feature graphs from random forests}
In the absence of available feature graphs mapping known relationships among features, such graphs can be constructed by observing how the model utilises features in the learning phase. A few graph-building approaches have been recently proposed. However, as for the methods presented in the previous section, they have been applied exclusively within supervised settings. 
Ruaud et al.~\cite{ruaud2022interpreting} computed the importance of individual features and their pairwise interactions in tree ensemble models based on the interaction effects between pairs of variables on the response, presenting them as a feature network with the primary aim of enhancing model interpretation. 
Other recent studies have constructed feature graphs directly from the structure of supervised random forests for various applications. In these studies, each tree is represented as a graph, where features represented are vertices and parent-child relations are edges. These studies also assume that graph representations of decision trees can offer concise, interpretable summaries of relationships among features. 

Kong et al.~\cite{kong2020forgenet} presented a graph extractor module that builds an unweighted directed feature graph from a supervised 
random forest based on the splitting order of features in decision trees. The graph extractor is added to the graph-embedded deep feed-forward network (GEDFN) model proposed in a previous work~\cite{kong2018graph}, which integrates the feature graph as a hidden layer within deep neural networks, enabling the use of GEDFN even when an existing feature graph is not available. The primary aim here was to achieve an informative sparse structure model for n$<<$p scenarios, as in omics data analysis.
Bayir et al.~\cite{bayir2022topological} converted each decision tree within a supervised forest model into a weighted directed feature graph and extracted key graph features to obtain a high-dimensional vector representation of each tree. Using TDA Mapper, they transformed these high-dimensional feature vectors into a topological cluster network of trees, with the goal of selecting representative decision trees for constructing smaller and more efficient random forests.
Additionally, Cantao et al.~\cite{cantao2022feature} represented random forest classifiers as weighted directed feature graphs and explored various centrality measures to rank features and perform effective feature selection. 

These studies construct feature graphs from supervised random forests, primarily trained for classification tasks, generally assigning unitary weight to each parent-child split and without considering leaf nodes. The resulting feature graphs are tailored to specific downstream tasks, such as feature selection, topological data analysis, or integration into neural architectures, without fully exploring their potential.
In contrast, our study focuses on building feature graphs from random forests in unsupervised settings, an area that remains largely unexplored. 
It proposes diverse edge-building criteria for constructing both general and cluster-specific feature graphs and presents various strategies for leveraging these graphs to underscore the potential of using feature graphs for both feature selection and model interpretability.


\section{Methods}\label{sec:MM}

\subsection{Unsupervised random forests}

Random Forests are ensembles of decision trees, with each tree denoted as $\tree$, constructed independently of each other using a bootstrapped or subsampled dataset. The training dataset comprises $n$ data points or samples $x_i$, with $i$ from 1 to $n$, over the feature space $\fs$, where $\fs = \{x^1, x^2, ..., x^d\}$ and $d$ is the dimensionality of the feature space, so that each $x_i$ is a vector from $R^d$, and $x_i^j$ denotes the value of feature $j$ for sample $i$.
%
Each tree $\tree$ represents a recursive partitioning of the feature space $\fs$ and each of its nodes $\node \in \tree$ corresponds to a subset, typically a hyper-rectangle, in the feature space $\fs$, denoted as $R(\node) \subset \fs$. The root node $\node_0$ is the initial node of the decision tree and corresponds to the entire feature space, i.e. $R(\node_0) = \fs$. The number of training samples falling into the hyper-rectangle $R(\node)$ is denoted as $\size(\node)$.

Each tree $\tree$ is grown using a recursive procedure which identifies two distinct subsets of nodes: the set of internal nodes $S(\tree)$ which apply a splitting rule to a partition of the feature space $\fs$, and the set of leaf nodes $L(\tree)$ which represent the terminal partitions of the feature space $\fs$ and, in supervised settings, provide prediction values for regression tasks or class labels for classification tasks.
A split at node $\node$ generates two children nodes $\node^\ch$ of node $\node$, a left child $\node^\lc$ and a right child $\node^\rc$, which divide $R(\node)$ into two separate hyper-rectangles, $R(\node^\lc)$ and $R(\node^\rc)$, respectively. 
The depth of node $\node$, denoted as $\depth(\node)$, is defined as the length of the path from the root node $\node_0$, so that $\depth(\node_0) = 0$ and  $\depth(\node^\lc) = \depth(\node^\rc) = \depth(\node) + 1 \;\; \forall \node \in \tree$.

The split at node $\node$ is decided in two steps. First, a subset $M(\node)$ of features is chosen uniformly at random. Then, the optimal split feature $x(\node) \in M(\node)$ and split value $z(\node) \in \R$ are determined by maximising the decrease in impurity given by:
\begin{equation}
\label{eq:impdecrease}
    \decrease(\node, x(\node), z(\node)) := \Impurity(\node) - \frac{\size(\node^\lc) \Impurity(\node^\lc) - \size(\node^\rc) \Impurity(\node^\rc)}{\size(\node)}
\end{equation}
where $\Impurity(\node)$ is some measure of impurity. 
Thus, $\decrease(\node, x(\node), z(\node))$ indicates the decrease in impurity for node $\node$ due to $x(\node)$ and $z(\node)$. 

In supervised contexts, impurity measures such as entropy and the Gini index are commonly utilised, leveraging the response vector $y$ to inform the splitting. In unsupervised scenarios, recent developments have introduced unsupervised splitting criteria, including a splitting rule inspired by the Fixation Index in population genetics~\cite{wright1949genetical,pfeifer2024federated}. 
The Fixation Index computes the average pairwise distances between samples within and across groups formed by a node split. Formally:

\begin{equation}
    \Delta_{\mathcal{F}}(\node, x(\node), z(\node)):=\frac{(\distance_{\square}(\node^{\lc}) + \distance_{\square}(\node^{\rc}))/2 }{\distance_{\nabla}(\node)}
\end{equation}
where
\begin{equation}
    \distance_{\square}(\node):= \frac{\sum_{i,h} (x_i(\node)-x_h(\node))^{2}}{\size(\node)(\size(\node)-1)}, \quad \text{for} \quad i\neq h
\end{equation}
and 
\begin{equation}
    \distance_{\nabla}(\node):= \frac{\sum_{i,h} (x_i(\node^{\lc})-x_h(\node^{\rc}))^{2}}{\size(\node^\lc)\size(\node^\rc)},
\end{equation}

where $x$ is a given candidate feature and $x_i,x_h$ refers to the specific value in sample $i$ and sample $h$. 
Once the unsupervised random forest is built, an affinity matrix is derived by counting the number of times each pair of samples appears in the same leaf node. The affinity matrix serves as input for distance-based clustering methods, such as Ward's linkage method~\cite{ward1963hierarchical,pfeifer2024federated}. 

The clustering algorithm generates a partition $\mathcal{C} = \{C_1, C_2, .., C_p\}$ of the data samples, where each $C_g$ (with $g$ ranging from $1$ to $p$) represents an individual cluster, and $p$ denotes the total number of identified clusters. For every sample $x_i$, the algorithm assigns a value $c_i$ denoting its membership to a cluster $C_g$. Consequently, any two samples $x_i$ and $x_h$ belong to the same cluster $C_g \in \mathcal{C}$ if and only if their assigned membership values are equal, i.e., $c_i = c_h$.

\subsection{Building the feature graph}

Consider a random forest $\forest$ trained on a dataset over the feature space $\fs$, comprising trees $\tree$ made of nodes $\node$, where each node is associated with a feature $x$ or a leaf.
A weighted directed graph $G$, denoted as $G(\forest) = (V, E, W)$, can be constructed from $\forest$, where:
\begin{itemize}
    \item $V$ is the set of vertices of the graph, representing the $d$ features in $\fs$ as well as $l$ to denote leaf nodes, i.e., $V = \{x^1, x^2, \dots, x^d\} \cup \{l\}$; 
    \item $E$ is the set of directed edges, where each edge is an ordered pair of vertices in $V$, i.e., $E = \{(\vertex_i, \vertex_j) \mid \vertex_i, \vertex_j \in V\}$;
    \item $W$ is the set of weights assigned to each directed edge, indicating some measure of capacity or cost associated with traversing the edge, i.e., $W = \{w_{ij} \mid (\vertex_i, \vertex_j) \in E\} $.
\end{itemize}

The graph can also be represented by its adjacency matrix $A(G)$, where $A[\vertex_i, \vertex_j] = w_{ij}$. 
Given these premises, the proposed method constructs a feature graph from the structure of the random forest with features as nodes, and edges reflecting the occurrence of any two features labelling adjacent nodes (parent and child nodes) 
across all trees of the forest. 
The elements of the adjacency matrix are then defined as follows:

\begin{equation}
\begin{split}
&A[x^i,x^j] = \sum_{\tree \in \forest}\sum_{\substack{\node \in S(\tree) \\ x(\node) = x^i \\ x(\node^\ch) = x^j}} q(\node, \node^\ch) \quad \forall \; x^i, x^j \in \fs;\\
&A[x^i,l] = \sum_{\tree \in F}\sum_{\substack{\node \in S(\tree) \\ x(\node) = x^i \\ \node^\ch \in L(\tree)}} q(\node, \node^\ch) \quad \forall \; x^i \in \fs;\\
&A[l,x^i] = 0 \quad \forall \; x^i \in \fs.
\end{split}
\end{equation}

where $q(\node, \node^\ch)$ is a quantity assigned to the node pair $\node$ and $\node^\ch$, defined differently according to one of the following four edge-building criteria. 
\begin{itemize}
    \item Under the \textit{present} criterion, $q(\node, \node^\ch) = 1$, thus counting the occurrences of any two features labelling adjacent nodes across all trees of the forest.
    \item According to the \textit{fixation} criterion, $q(\node, \node^\ch) = \Delta_{\mathcal{F}}(\node, x(\node), z(\node))$, where the contribution of each edge corresponds to the fixation index computed at that node, prioritising more effective splits.
    \item Under the \textit{level} criterion, $q(\node, \node^\ch) = \depth(\node^\ch)^{-1}$, scaling the contribution of each edge by the depth of the node, giving more weight to splits closer to the root, which are assumed to have a greater impact on the outcome by affecting more samples.
    \item As per the \textit{sample} criterion, $q(\node, \node^\ch) = \frac{N(\node^\ch)}{N(\node_0)}$, 
    with the contribution of each edge scaled by the number of data samples traversing that edge over the total number of samples, emphasising splits that affect more samples in the dataset. 
\end{itemize}

The proposed approach can be tailored to construct cluster-specific feature graphs by scaling the edge weights (determined based on a specified criterion) by the proportion of data samples associated with a particular cluster assignment. This adjustment assigns greater importance to splits that predominantly affect samples assigned to a given cluster, under the assumption that if certain features effectively discriminate a particular cluster, samples assigned to that cluster will predominantly traverse paths in the tree containing those features. 

To construct a cluster-specific graph, each quantity $q(\node, \node^\ch)$ is multiplied by a factor $f(\node, \node^\ch, C)$, defined as:
\begin{equation}
f(\node, \node^\ch, C) = \frac{|\{x_i \in \node^\ch \land c_i = C\}|}{N(\node^\ch)}
\end{equation}
where the numerator of the fraction computes the cardinality of the set of samples traversing the edge and assigned to a given cluster $C$. Notably, for every $q(\node, \node^\ch)$, the cluster-specific multiplicative factors $f(\node, \node^\ch, C)$ sum to 1, i.e.,

\[ \sum_{C \in \mathcal{C}} f(\node, \node^\ch, C) = 1 \;\; \forall \; \node, \node^\ch \in t,\; \forall \; \tree \in \forest\]

Consequently, the constructed feature graph can be regarded as the sum of its cluster-specific feature graphs.

\subsection{Mining the feature graph}\label{sec:MG}

The feature graph built on the structure of a random forest trained in an unsupervised manner can be used to identify the features most effective in splitting data into clusters. 
In a weighted directed graph, the influence of a vertex can be measured using a variety of centrality measures, and out-degree centrality in particular. The weighted out-degree centrality $C_{\text{out}}^{\text{weighted}}(\vertex_i)$ of a vertex $\vertex_i$ is the sum of the weights of edges outgoing from $\vertex_i$:
\[ C_{\text{out}}^{\text{weighted}}(\vertex_i) = \sum_{\vertex_j \in V} w_{ij} \]


The constructed graph can also be leveraged to identify subsets of relevant features for feature selection purposes. 
To this aim, a brute-force and a greedy approach are proposed to identify subsets of vertices connected by heavy weights in the graph. Both strategies consider edges connecting features while excluding $l$ and its incident edges, as it represents terminal tree nodes not associated with any splitting feature. Additionally, they do not take into account self-edges or the direction of the edges. In this context, the weighted undirected graph  $G_\fs = (V_\fs, E_\fs, W_\fs)$ is defined, induced by the feature set $\{x^1, x^2, \dots, x^d\}$, with edge weights equal to the average of the weights of corresponding edges in the directed graph, i.e., $V_\fs = \fs$, $E_\fs = \{\{\vertex_i, \vertex_j\} \mid \vertex_i,\vertex_j \in V_\fs, \vertex_i \neq \vertex_j, (\vertex_i, \vertex_j) \in E\}$, and $W_\fs = \{ \frac{w_{ij} + w_{ji}}{2}  \mid \vertex_i,\vertex_j \in V_\fs, (\vertex_i, \vertex_j) \in E, w_{ij} \in W \}$.

\subsubsection{Brute-force feature selection}

The top $k$ features can be evaluated in a brute-force manner by inspecting all connected subgraphs of size $k$ in $G_\fs$. 

A subgraph $G' = (V', E', W')$ of size $k$ is defined as the subgraph of $G_\fs$ induced by the set of $k$ vertices $V' \subseteq V_\fs$, containing all edges in $ E_\fs $ whose endpoints are in $ V' $, i.e. $E' = \{\{\vertex_i, \vertex_j\} \in E_\fs \mid \vertex_i, \vertex_j \in V' \}$.
The resulting graph $G'$ is said to be connected if, for every pair of vertices $\vertex_i, \vertex_j \in V'$, there exists a path in the graph from vertex $\vertex_i$ to vertex $\vertex_j$. Formally, $G'$ is connected if and only if there is a sequence of vertices $\vertex_1, \vertex_2, \ldots, \vertex_k \in V'$ such that for each $i$ with $1 \leq i < k$, the edge $\{\vertex_i, \vertex_{i+1}\} \in E'$ exists.

The total weight (TW) of the subgraph $G'$ is given by the sum of weights of all edges in $ E' $, and the average weight (AW) can be defined as the total weight divided by the maximum number of edges, i.e.,
\[
\text{TW}(G') = \sum_{\{\vertex_i, \vertex_j\} \in E'} w_{ij} \in W' \;\;\;\;
\text{AW}(G') = \frac{2{TW}(G')}{|V'|(|V'|-1)}
\]

The top $k$ features according to the proposed brute-force approach correspond to the vertex set $V' \subseteq V_\fs$ of size $k$ inducing a subgraph $G'$ of $G_\fs$ that maximises the average (or total) edge weight, i.e.,
\[
V' = \arg \max_{\substack{V' \subseteq V_\fs \\ |V'| = k \\ G' \text{ is connected}}} \text{AW}(G')
\]

This approach faces computational limitations due to the growth in the number of possible subgraphs relative to the size of the feature set. Specifically, for $d$ features, the maximum number of subgraphs of size $k$ is determined by $\frac{d!}{k!(d-k)!}$. Consequently, this method exhibits exponential computational complexity and becomes infeasible for large feature spaces. 

\subsubsection{Greedy feature selection}
The top $k$ features can be determined in a greedy manner by initially selecting the two features associated with vertices in $G_\fs$ that are connected by the heaviest edge. Subsequently, features are iteratively added by maximising the average weights of edges with the already selected features. Given a graph $G_\fs$, a vertex $\vertex_i \in V_\fs$ and a vertex set $V' \subset V_\fs$ such that $\vertex_i \notin V'$, the average weight of the new edges (AWN) connecting $\vertex$ and $V'$ can be defined as: 
\[
\text{AWN}(G_\fs, \vertex_i, V') =\frac{1}{|V'|}\sum_{\vertex_j \in V'} w_{ij} \in W_\fs
\]
The greedy procedure outlined in Algorithm~\ref{alg:greedy} yields the top $k$ features from the graph $G_\fs$, along with two weight lists: the average edge weight of the subgraph induced by the selected feature set and the average edge weight connecting the newly added feature to the previously selected features. These weight lists offer valuable insights into determining the optimal number of features. The algorithm operates by evaluating all neighbours of selected nodes at each iteration, resulting in approximately $k*|E_\fs|$ operations, which approaches $k*d^2$ operations for dense graphs (which feature graphs often are). Consequently, this method exhibits polynomial computational complexity and can be efficiently implemented.


\begin{algorithm}
\small
\caption{Top $k$ greedy feature selection algorithm}
\label{alg:greedy}
\begin{algorithmic}[1]
\STATE \textbf{Input:} $G_\fs = (V_\fs, E_\fs, W_\fs)$, $k$
\STATE \textbf{Output:} $V'$
\STATE \textbf{Initialise:} $\added \gets \emptyset$, $\toadd \gets V_\fs$
\hfill\COMMENT{initialise added and to-add feature sets}
\STATE \textbf{Initialise:} $avg_n \gets [\;]$, $avg \gets [\;]$
\hfill\COMMENT{initialise lists for edge weight sum and avg}
\STATE $\{\vertex_1, \vertex_2\} = \arg \max_{\vertex_i,\vertex_j \in G_\fs} w_{ij} \in W_\fs$ 
\hfill\COMMENT{find vertices with max edge weight}
\STATE $avg$.append($w_{ij}$), 
       $avg_n$.append($w_{ij}$) 
\hfill\COMMENT{update edge weight lists}
\STATE $\added \gets \{\vertex_1, \vertex_2\} \cup \added$, $\toadd \gets \toadd \setminus \{\vertex_1, \vertex_2\}$
\hfill\COMMENT{update feature sets}
\WHILE{$|\added| <= k$}
    \STATE $\vertex_i = \arg \max_{\vertex \in \toadd} \text{ AWN}(G_\fs, \vertex, \added)$
    \hfill\COMMENT{find vertex with max avg edge weight}
    \STATE $avg$.append($\text{AW}(G_\fs[\added])$) 
    \hfill\COMMENT{update edge weight lists}
    \STATE $avg_n$.append($\text{AWN}(G_\fs, \vertex_i, \added)$) 
    \STATE $\added \gets \{\vertex_i\} \cup \added$, $\toadd \gets \toadd \setminus \{\vertex_i\}$
    \hfill\COMMENT{update feature sets}
\ENDWHILE
\RETURN{$\added$, $avg$, $avg_n$}
\end{algorithmic}
\end{algorithm}

For both feature selection strategies, all features are eligible for selection only if the constructed feature graph is connected. The probability of a graph being connected depends on the density of the graph, defined as the ratio of actual edges to possible edges, i.e., equivalent to $\frac{2|E_\fs|}{(|V_\fs|)(|V_\fs| - 1)}$ for undirected graphs and $\frac{|E_\fs|}{(|V_\fs|)(|V_\fs| - 1)}$ for directed graphs. As the density of edges increases, the probability of finding a path between any pair of vertices typically increases. In random graphs, there often exists a critical edge density above which the graph is almost surely connected~\cite{bollobas1998random}. The density of the feature graph is thus determined by the number of features ($|V_\fs|$) and the number of edges ($|E_\fs|$), which depend on the number of trees. To achieve a connected feature graph it is often sufficient to train a greater number of trees. Alternatively, both feature selection strategies can be modified so that, initially, the connected components of the graph are identified using a graph traversal algorithm. Then, the feature selection approach can be deployed concurrently in the largest connected components. In summary, while the proposed approaches rely on a connected feature graph, this assumption is often satisfied and generally attainable by training a greater number of trees. In the event of a disconnected graph, the algorithms can be adapted to operate on the largest connected components.

\section{Evaluation}\label{sec:EV}


In this evaluation section, we conduct a comprehensive analysis of the methodologies outlined in \Cref{sec:MM}. Initially, we assess the effectiveness of the proposed graph construction methods, which encompass four edge-building criteria. This evaluation entails a comparison across synthetic datasets, examining the correlation between node centrality and feature relevance, as well as the relationship between edge weight and the discriminatory power of feature pairs. While \Cref{sec:EV1} analyses feature graphs constructed on the entire datasets, \Cref{sec:EV2} focuses on cluster-specific feature graphs. Subsequently, we evaluate the proposed brute-force and greedy feature selection methods across synthetic datasets. These datasets comprise varying numbers of relevant and irrelevant features in \Cref{sec:EV3}, along with redundant features in \Cref{sec:EV4}.
Furthermore, we evaluate the efficacy of the greedy feature selection method for dimensionality reduction purposes on benchmark datasets in \Cref{sec:EV5}. Finally, we demonstrate the utility of greedy feature selection for interpretability on an omics dataset in a disease subtyping application in \Cref{sec:EV6}.

\subsection{Out-degree centrality and edge weight}\label{sec:EV1}

The initial evaluation aims to assess whether the generated feature graphs exhibit two desirable properties:
(i) features that are more relevant in discriminating clusters should exhibit higher out-degree centrality in the graph, indicating that out-degree centrality effectively captures the relative importance of features;
(ii) pairs of features that, used together, are more effective in separating clusters should be connected by edges with greater weight, showing that edge weight accurately reflects the ability of the feature pair to discriminate clusters. 

\begin{figure}[H]%
    \centering
    \includegraphics[width=\textwidth]{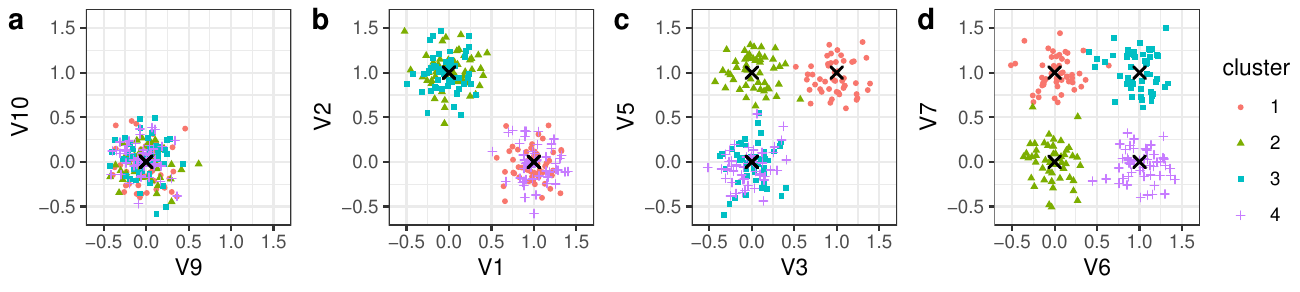}
    \caption{Feature pairs discriminating (a) 1, (b) 2, (c) 3 and (d) 4 clusters.}
    \label{fig:v1clusters}
    \medskip
\end{figure}

To evaluate these properties across the four edge-building criteria, two sets of synthetic datasets were generated. 
First, 30 synthetic datasets, each comprising 13 features (3 relevant and 10 irrelevant), were generated to form four clusters. In each cluster, 50 data points were distributed normally around predefined cluster centres, with a standard deviation of 0.2 to ensure well-defined clusters. For relevant features, data points were generated around distinct cluster centres ([1,0,0,0] for $V1$, [0,1,0,0] for $V2$, and [0,0,1,0] for $V3$), while for irrelevant features, data points were generated around the same cluster centres [0,0,0,0]. Then, 30 synthetic datasets of 13 features, 8 relevant and 5 irrelevant, were generated around four cluster centres so that different feature pairs could distinguish a varying number of clusters ranging from 1 to 4. 
For instance, values for features $V3$ and $V4$ were generated around cluster centres [1,0,0,0] and [0,1,1,1], so two globular clusters can be distinguished in the two-dimensional space defined by the two features, centred in (0,1) and (1,0). \Cref{fig:v1clusters} illustrates how data was generated so that different feature pairs had different discriminating power. 

For these and all experiments conducted in this manuscript, random forests were trained in an unsupervised manner using the fixation index as a splitting rule, employing the uRF library~\cite{pfeifer2024federated}. The size of leaf nodes was set to 5, and the number of features sampled uniformly at random at each node was determined as the square root of the dimensionality of the feature space. Feature graphs were generated according to the four proposed edge-building criteria and evaluated in terms of vertex centralities and edge weights.

\subsection{Cluster-specific feature graphs}\label{sec:EV2}


The subsequent evaluation aims to determine whether the generated cluster-specific feature graphs can effectively capture the role of each feature in discriminating individual clusters. Initially, a collection of 30 synthetic datasets, each containing 13 features, with 4 of them relevant and 9 irrelevant, was generated around four cluster centres. Data for the four relevant features were generated as detailed in Sec.~\ref{sec:EV1}, with data points distributed around the cluster centres [1,0,0,0], [0,1,0,0], [0,0,1,0], and [0,0,0,1]. This configuration ensured that each relevant feature could contribute to the discrimination of all clusters, while individually, each feature could effectively distinguish one cluster from all others. Subsequently, cluster-specific feature graphs were constructed for each cluster from forests trained on these datasets. Within each cluster-specific graph, the comparison of out-degree centrality was conducted on features categorised into three groups: cluster-specific (the feature capable of singularly identifying the cluster), sub-relevant(features collectively contributing to identifying the cluster), and irrelevant (features unable to discriminate the cluster).

\subsection{Feature selection on synthetic datasets with relevant features}\label{sec:EV3}
The proposed brute-force and greedy selection strategies were assessed on their ability to detect a varying number of relevant and irrelevant features. Synthetic datasets consisting of 13 features, with $q$ relevant features ranging from 3 to 7, were generated around $q+1$ predefined cluster centres. The cluster centres were defined so that the $i^{th}$ relevant feature generated data normally distributed around 1 for the $(i+1)^{th}$ cluster and 0 for the others. For instance, in the case of 3 relevant features, cluster centres were set as $[0,1,0,0]$ for $V1$, $[0,0,1,0]$ for $V2$, and $[0,0,0,1]$ for $V3$. This configuration ensured that all relevant features were necessary and equally important in differentiating clusters. For each configuration, 30 datasets were generated. Feature graphs were constructed from random forests trained on these datasets using the four edge-building criteria. The brute-force and greedy methods were then employed to select the top $k$ features, with $k$ ranging from 2 to 12, while recording the average weight of the subgraph induced by the selected features. The percentage of relevant features among the selected features was computed for each feature graph.


\subsection{Feature selection on synthetic datasets with repetitive features}\label{sec:EV4}
The proposed feature selection strategies for mining the constructed feature graphs were then evaluated on their ability to handle redundant features, by identifying combinations of features that provide complementary information. To validate this claim, synthetic datasets of 10 features, with the first 6 being relevant and the remaining 4 irrelevant, were generated around 4 cluster centres. In these datasets, pairs of relevant features were generated around identical cluster centres, thus providing overlapping information ([1,0,0,0] for V1 and V2, [0,1,0,0] for V3 and V4, [0,0,1,0] for V5 and V6). Consequently, any set of three features including either V1 or V2, V3 or V4, and V5 or V6 (amounting to 8 out of 120 possible feature combinations) can effectively distinguish all clusters. Feature graphs were constructed from random forests trained on these datasets according to the \textit{sample} criteria. The average edge weight of every triad in the graph was evaluated to detect optimal feature combinations.
 
\subsection{Feature selection on benchmark datasets}\label{sec:EV5}

The proposed greedy feature selection approach, here denoted as the \textit{graph-based} method, was compared against an alternative tree-based strategy, referred to as \textit{impurity-based}. The \textit{impurity-based} method utilises clustering predictions from an unsupervised forest to train a forest in a supervised manner using these predictions as the response variable while assessing feature importance based on impurity criteria. Features were subsequently ranked in descending order of impurity, and those with the highest impurity-based feature importance were selected. The performance of the two feature selection approaches was then compared by evaluating the quality of clustering predictions made by unsupervised random forests trained on the top $k$ features selected by each method. The performance of Ward clustering on affinity matrices derived from the unsupervised random forests was quantified using the Adjusted Rand Index (ARI). Importantly, it should be noted that both feature selection strategies are unsupervised, as neither leverages ground truth information during model training.

The two feature selection approaches were evaluated on the 10 benchmark datasets detailed in Table~\ref{tab:Datasets}. For each dataset, random forests consisting of 1000 trees were trained 30 times in an unsupervised manner to generate clustering predictions. Feature graphs were constructed from the structure of the forest for each iteration and then averaged to produce a final feature graph for the dataset. Feature importance was determined by computing the out-degree centrality of each feature in the graph.
The clustering predictions were utilised to train 30 random forests of 1000 trees in a supervised manner, employing Gini impurity as the measure. The feature importance for each supervised forest was computed using impurity-corrected criteria~\cite{nembrini2018revival}, and the final feature importance was averaged across the 30 forests. The correlation between the feature importance obtained from the unsupervised and supervised approaches was evaluated using Pearson's correlation coefficient.
Subsequently, the top $k$ features were identified using the two methods: applying the greedy feature selection algorithm to the feature graph, as per \textit{graph-based feature selection}, and selecting the $k$ features with the highest Gini impurity scores, according to \textit{impurity-based feature selection}. 
Then, 30 random forests of 500 trees were trained for clustering on the top $k$ features, with $k$ ranging from 2 to the total number of features, and the quality of the clustering solutions was evaluated by computing the ARI.



\begin{table*}[h!]
  \begin{center}
    \caption{Benchmark Datasets}
    \label{tab:Datasets}
    \begin{tabular}{l c c c c} 
     &  &  &  & \#Samples per \\ 
      \textbf{Datasets} & \#Samples & \#Features & \#Cluster & Cluster \\ 
      \hline
      Iris          & 150 & 4 & 3  & 50,50,50\\
      Liver         & 345 & 6 & 2  & 145,200 \\ 
      Ecoli         & 336 & 7 & 8  & 143,77,52,35,20,5,2,2\\
      Breast cancer & 106 & 9 & 6  & 21,15,18,16,14,22\\
      Glass         & 214 & 9 & 4  & 70,76,17,51\\
      Wine          & 178 & 13 & 3 & 59,71,48 \\
      Lymphography  & 148 & 18 & 4 & 2,81,61,4 \\
      Parkinson     & 195 & 22 & 2 & 48,147 \\
      Ionosphere    & 351 & 34 & 2 & 225,126 \\ 
      Sonar         & 208 & 60 & 2 & 97,111 \\
      \hline
    \end{tabular}
  \end{center}
  \centering
\end{table*}

\subsection{Interpretable disease subtype discovery}\label{sec:EV6}


The graph-building and graph-mining methods proposed in this study were applied to disease subtyping. Unsupervised random forests were utilised to derive an affinity matrix from patients diagnosed with kidney cancer, which served as input for a clustering method. Subsequently, the inferred clusters were examined using the proposed methodologies, focusing on genes and gene interactions and their relevance to each patient cluster. Specifically, we demonstrate the practical utility of our approaches using gene expression data obtained from patients suffering from kidney cancer, as provided by~\cite{rappoport2018multi}. The initial dimensions of the retrieved data table consisted of 606 patients and 20,531 genes. The obtained gene expression patterns were filtered by selecting the top 100 features with the highest variance and the bottom 100 features with the lowest variance. Following this pre-processing step we identified the top-15 relevant genes using the herein proposed greedy selection strategy (see Section~\ref{sec:MG}). Based on this gene subset, we performed disease subtyping employing unsupervised random forests as introduced by Pfeifer et al.~\cite{pfeifer2024federated}. We then drew survival curves for the identified clusters and assessed differences using the Cox log-rank test, an inferential procedure for comparing event time distributions among independent (i.e., clustered) patient groups. Finally, we presented findings on genes, gene interactions, and their importance to each patient cluster.


\section{Results and discussion}\label{sec:RD}

\subsection{Evaluation on synthetic datasets}

\subsubsection{Out-degree centrality and edge weight}
\begin{figure}[H]%
    \centering
    \begin{subfigure}[b]{0.38\textwidth}
        \centering
        \includegraphics[width=\textwidth]{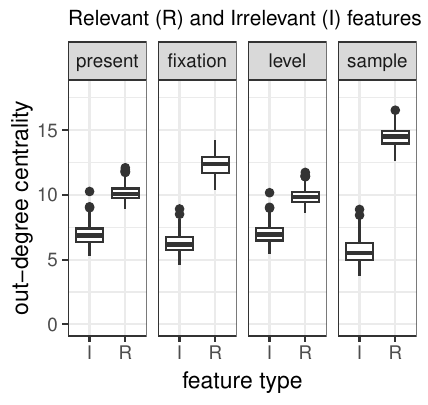}
        \caption{feature centrality}
    \end{subfigure}\hfil
    \begin{subfigure}[b]{0.6\textwidth}
        \centering
        \includegraphics[width=\textwidth]{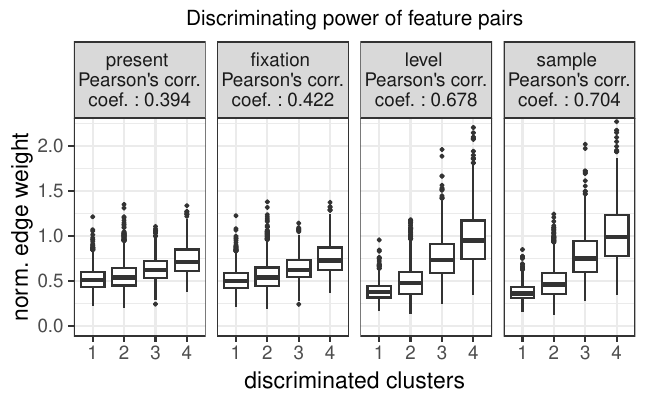}
        \caption{feature pair edge weights}
    \end{subfigure}\hfil
    \caption{ Evaluating out-degree centrality and edge weights of the feature graphs. }  
    \label{fig:v1}
\end{figure}

The first experiment detailed in Section~\ref{sec:EV1} reveals promising attributes of the constructed feature graph in terms of centrality and edge weight. Across all four edge-building criteria, the centrality of relevant features is consistently greater than that of irrelevant features, as evidenced in \Cref{fig:v1}~(a). T-tests with a significance threshold of 0.05 confirm that these differences are statistically significant ($p$-value $<$ e-16). Among all proposed criteria, the \textit{sample} criterion emerges as the most effective in discriminating relevant and irrelevant features, followed by \textit{fixation}. This suggests that the chosen centrality metric can effectively capture the importance of features in solving a clustering task.

The second experiment outlined in Section~\ref{sec:EV1} investigates the correlation between the weight of an edge connecting any two features and the ability of those features to separate data into clusters, as illustrated in \Cref{fig:v1}~(b). Across all edge-building criteria, there exists a significant positive correlation between the edge weight and the number of separated clusters, as quantified by Pearson's correlation coefficient. Once again, the \textit{sample} criterion stands out as the most effective, recording the highest correlation coefficient, followed by \textit{level}.  Consequently, the edge weight between two features emerges as a reliable indicator of their effectiveness in cluster separation, suggesting that feature selection strategies should prioritise features connected by edges with heavy weights. These considerations motivate the proposed feature selection strategies.


\subsubsection{Cluster-specific feature graphs}
\begin{figure}[H]%
    \centering
    \includegraphics[width=\textwidth]{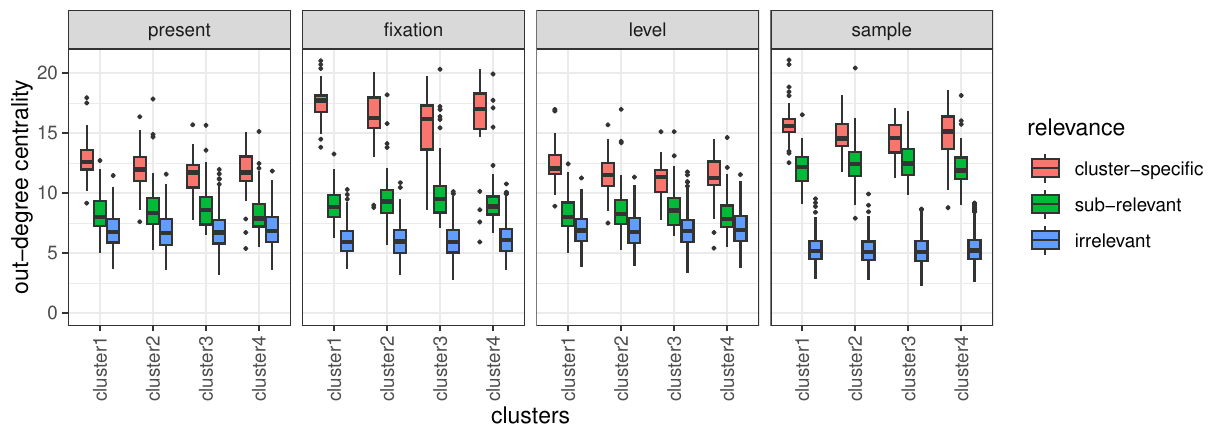}
    \caption{ Evaluating cluster-specific feature graphs.}
    \label{fig:v2}
    \medskip
\end{figure}

Further experiments on synthetic datasets reveal that the proposed cluster-specific scaling factor, used in conjunction with one of the defined edge-building criteria, can generate feature graphs able to discriminate between cluster-specific features, sub-relevant features as well as irrelevant features.
For every cluster-specific graph and under each edge-building criterion, the out-degree of the cluster-specific features exceeded that of sub-relevant features, which, in turn, was greater than that of irrelevant features, as illustrated in \Cref{fig:v2}. T-test confirmed the differences between the three groups of features to be significant for each examined criterion ($p$-value $<$ e-07).
The fixation criterion emerged as the most adept at isolating the cluster-specific feature, while the sample approach proved most effective at distinguishing between relevant and irrelevant features. 
The proposed approach presents a promising strategy for evaluating cluster-specific feature importance. Its inherent additive nature, which generates cluster-specific graphs that sum to the overall feature graph, also allows to easily compute feature graphs for relevant sets of clusters.

\subsubsection{Feature selection on synthetic datasets with relevant features}
\begin{figure}[H]%
    \centering
    \includegraphics[width=\textwidth]{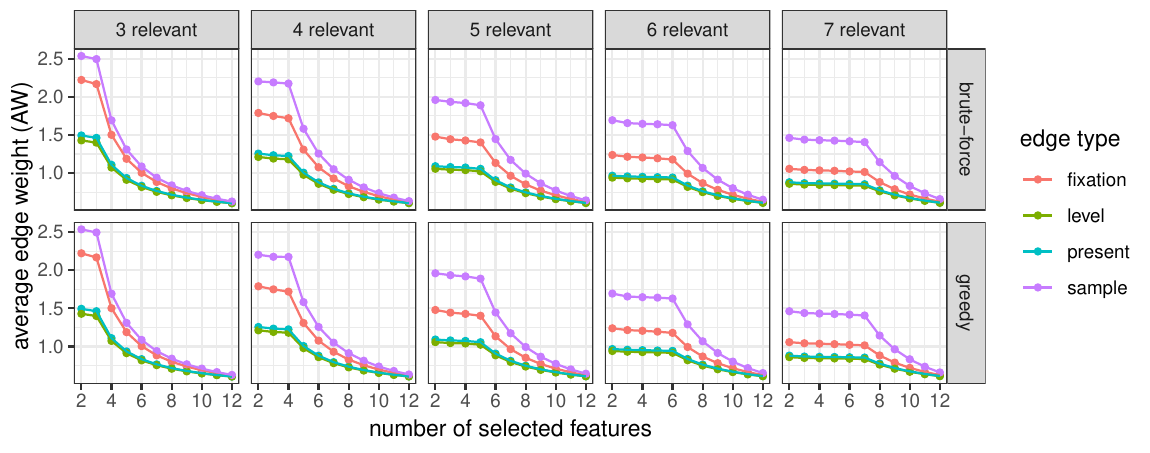}
    \caption{ Evaluating brute force and greedy feature selection approaches.}
    \label{fig:v3}
    \medskip
\end{figure}

Two proposed strategies for mining the constructed feature graphs, a brute force and a greedy approach, both focused on maximising the weight of edges connecting selected features, show promise in identifying the top $k$ features in a clustering task. In experiments on feature graphs generated from synthetic datasets with a varying number of relevant features, both approaches consistently selected all relevant features before any irrelevant ones demonstrating comparable performance, although the brute force approach has exponential computational complexity. 
Evaluating the average weight of the subgraph induced by the selected features can also provide valuable insights on the optimal number of features. In both graph mining methods, the inclusion of an irrelevant feature reduces the average weight of the subgraph because relevant features typically connect through heavier edges, whereas irrelevant ones participate in weaker connections. Hence, a notable decrease in the average weight of the subgraph can indicate the inclusion of an irrelevant feature, suggesting that the optimal number of features has been reached. 
This is supported by results in \Cref{fig:v3}, where a noticeable decline in the average edge weight occurs right after selecting all relevant features, especially under \textit{sample} or \textit{fixation} edge-building criteria.

\subsubsection{Feature selection on synthetic datasets with repetitive features}
\begin{figure}[H]%
    \centering
    \begin{subfigure}[b]{0.31\textwidth}
        \centering
        \includegraphics[width=\textwidth]{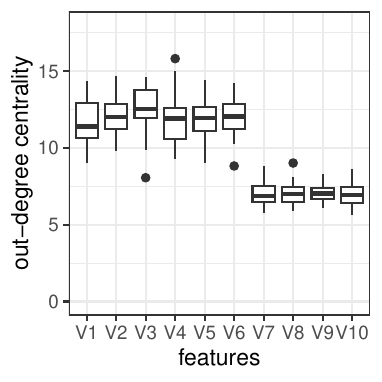}
        \caption{centrality}
    \end{subfigure}\hfil
    \begin{subfigure}[b]{0.68\textwidth}
        \centering
        \includegraphics[width=\textwidth]{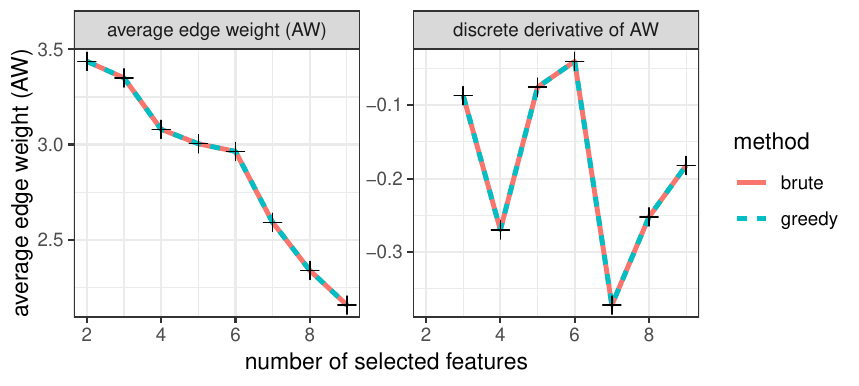}
        \caption{AW of subgraph of increasing size with derivative}
    \end{subfigure}\hfil
    \medskip
        \begin{subfigure}[b]{\textwidth}
        \centering
        \includegraphics[width=0.98\textwidth]{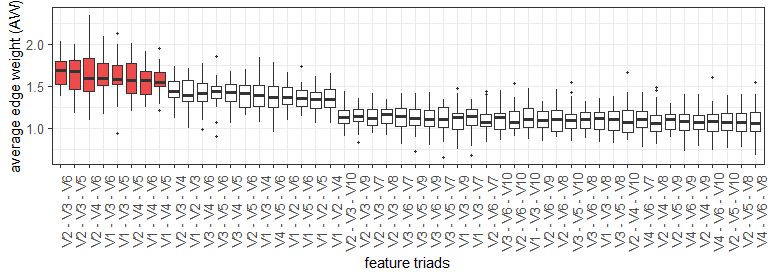}
        \caption{average edge weight of top 50 triads (8 effective feature combinations shown in red).}
    \end{subfigure}\hfil
    \caption{Evaluating feature selection strategies in case of redundant features. 
    }  
    \label{fig:v4}
\end{figure}

An alternative approach for feature selection involves ranking the top $k$ features by their out-degree centrality. 
However, experiments conducted on synthetic data with redundant features show that the out-degree centrality is comparable for all relevant features, as illustrated in \Cref{fig:v4}~(a), and alone it does not assist in identifying effective feature combinations. Conversely, graph mining strategies evaluating the edge weight of subgraphs enable a more effective selection. 
Across all synthetic datasets, the heaviest triad consistently corresponded to one of the effective feature combinations, and the eight heaviest triads, on average, were exactly the eight correct feature combinations, as shown in \Cref{fig:v4}~(c).
%
When the number of features to select is unknown, exploring the graph structure with brute force and greedy approaches proves insightful. The two methods select the same features and experience two substantial drops in the average edge weight of the expanding feature set, as depicted in \Cref{fig:v4}~(b): the first after including three features, the minimum required to discriminate all clusters, and the second after reaching six features, the total number of relevant features. 
%

\subsection{Evaluation on benchmark datasets}
\begin{figure}
    \centering
    \includegraphics[width=\textwidth]{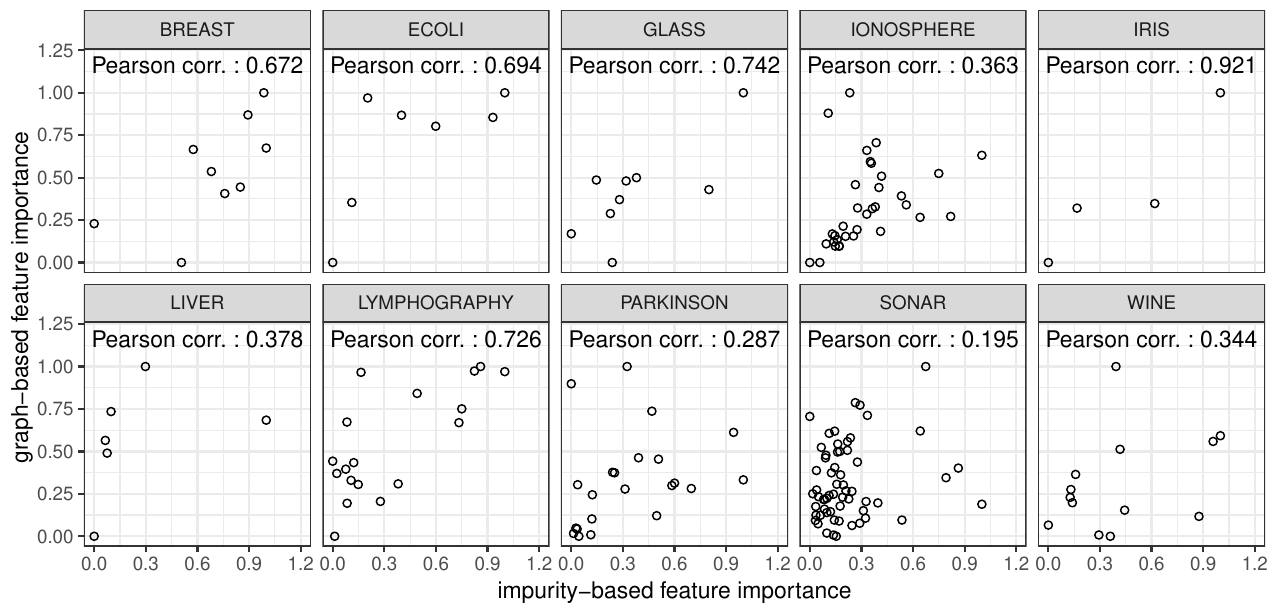}
    \caption{Correlation between \textit{graph-based} and \textit{impurity-based} feature importance.}
    \label{fig:v51}
    \medskip
\end{figure}

The effectiveness of the proposed methods in estimating feature importance and facilitating feature selection was evaluated on clustering (classification) benchmarks. 
Results in Figure \ref{fig:v51} show a partial agreement between the feature importance computed from the feature graph as out-degree centrality and the impurity-based feature importance derived from supervised random forests. Pearson's correlation has positive coefficients across all datasets, with significant p-values for ``Breast'', ``Glass'', ``Ionosphere'', and ``Lymphography''. 
%
Unsupervised random forests trained on the top $k$ features identified by the proposed greedy approach, and the top $k$ features based on impurity-driven feature importance show that clustering performance is comparable for smaller datasets  (which consist of a small set of curated features), while it is consistently superior with greedy selection for larger datasets (possibly including irrelevant features). Improved performance is particularly pronounced in datasets such as ``Lymphography'', ``Parkinson'', and ``Sonar''.
Furthermore, it is notable that \textit{graph-based feature selection} exhibits a higher degree of monotonicity in performance. Specifically, clustering performance tends to increase steadily with the number of features. In contrast, performance with \textit{impurity-based feature selection} fluctuates with the number of features, as evident in the ``Breast'' dataset. Establishing the optimal number of features is often challenging, and arbitrary cutoffs are commonly applied. The observed monotonicity is an added benefit of the proposed approach, as it mitigates the negative impact of selecting a suboptimal number of features. 

\begin{figure}
    \centering
    \begin{subfigure}[b]{0.61\textwidth}
        \centering
        \includegraphics[width=\textwidth]{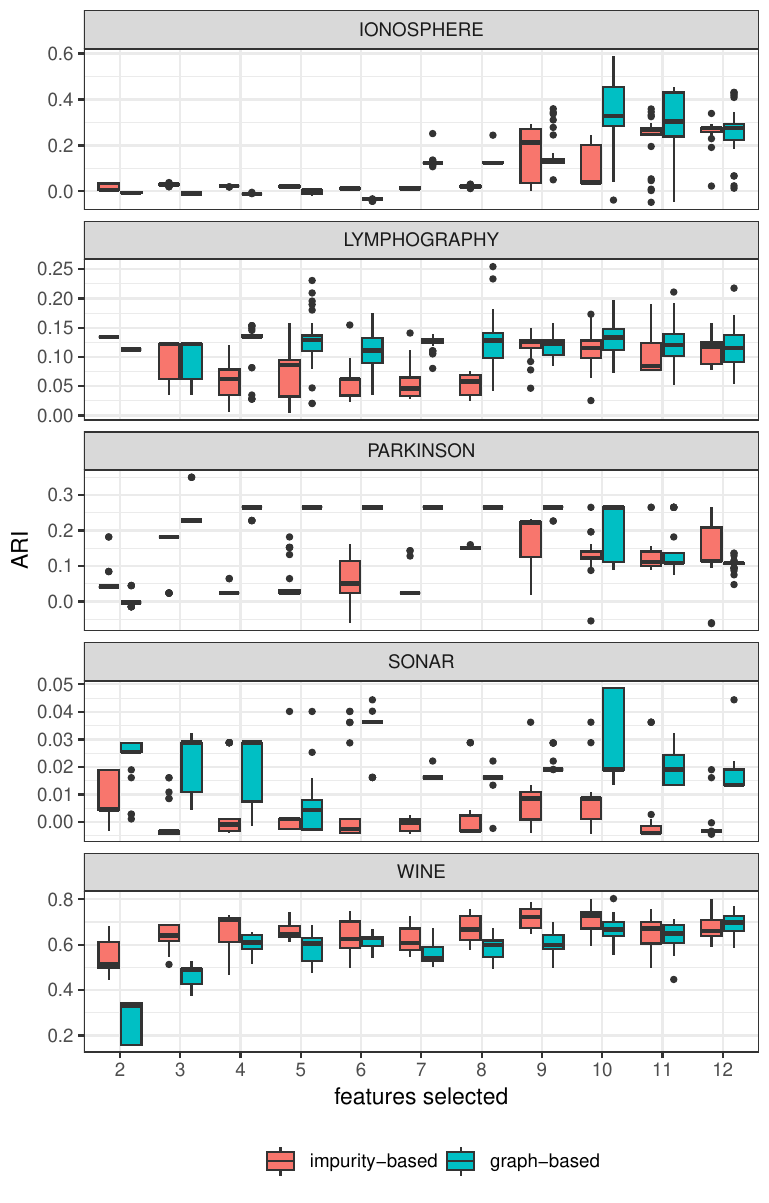}
        \caption{larger datasets}
    \end{subfigure}\hfil
    \begin{subfigure}[b]{0.36\textwidth}
        \centering
        \includegraphics[width=\textwidth]{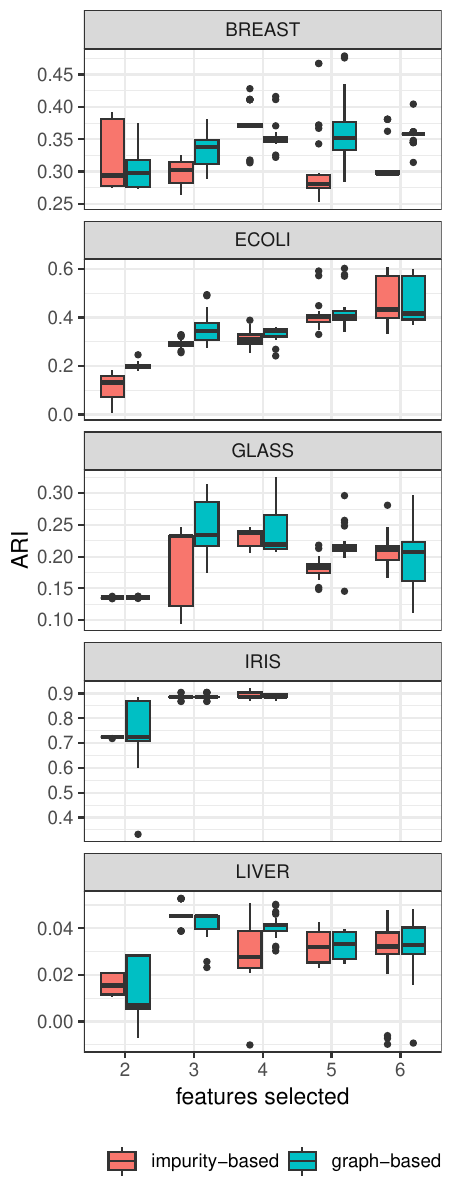}
        \caption{smaller datasets}
    \end{subfigure}\hfil
    \label{fig:v52}
    \caption{
Clustering performance, evaluated as Adjusted Rand Index (ARI), of unsupervised random forests constructed on $k$ selected features, with $k$ ranging from 2 to 12 for larger datasets and from 2 to 6 for smaller datasets. Features are selected according to \textit{graph-based} and \textit{impurity-based} feature selection.}
\end{figure}%


\subsection{Disease subtype discovery}


Disease subtyping via clustering employs computational methods to group patients based on shared characteristics, such as clinical features, genetic profiles, or biomarker expression patterns, enabling a deeper understanding of disease complexity and offering potential for personalised medicine and improved patient outcomes~\cite{rappoport2018multi,pfeifer2023parea}.
Understanding the varying importance of specific biomarkers and genes within risk clusters is crucial, underscoring the need for interpretability. Certain clusters may exhibit a higher relevance of particular genes or biomarkers, indicating their significance in understanding the risk associated with those subgroups. Identifying and prioritising these distinctive genetic factors within clusters is essential for devising targeted interventions and conducting personalised risk assessments. However, cluster interpretability is not fully addressed by previous work. The methods outlined in this work aim to bridge this gap by offering enhanced interpretability for unsupervised learning.

\begin{figure}
    \centering
    \includegraphics[width=1\textwidth]{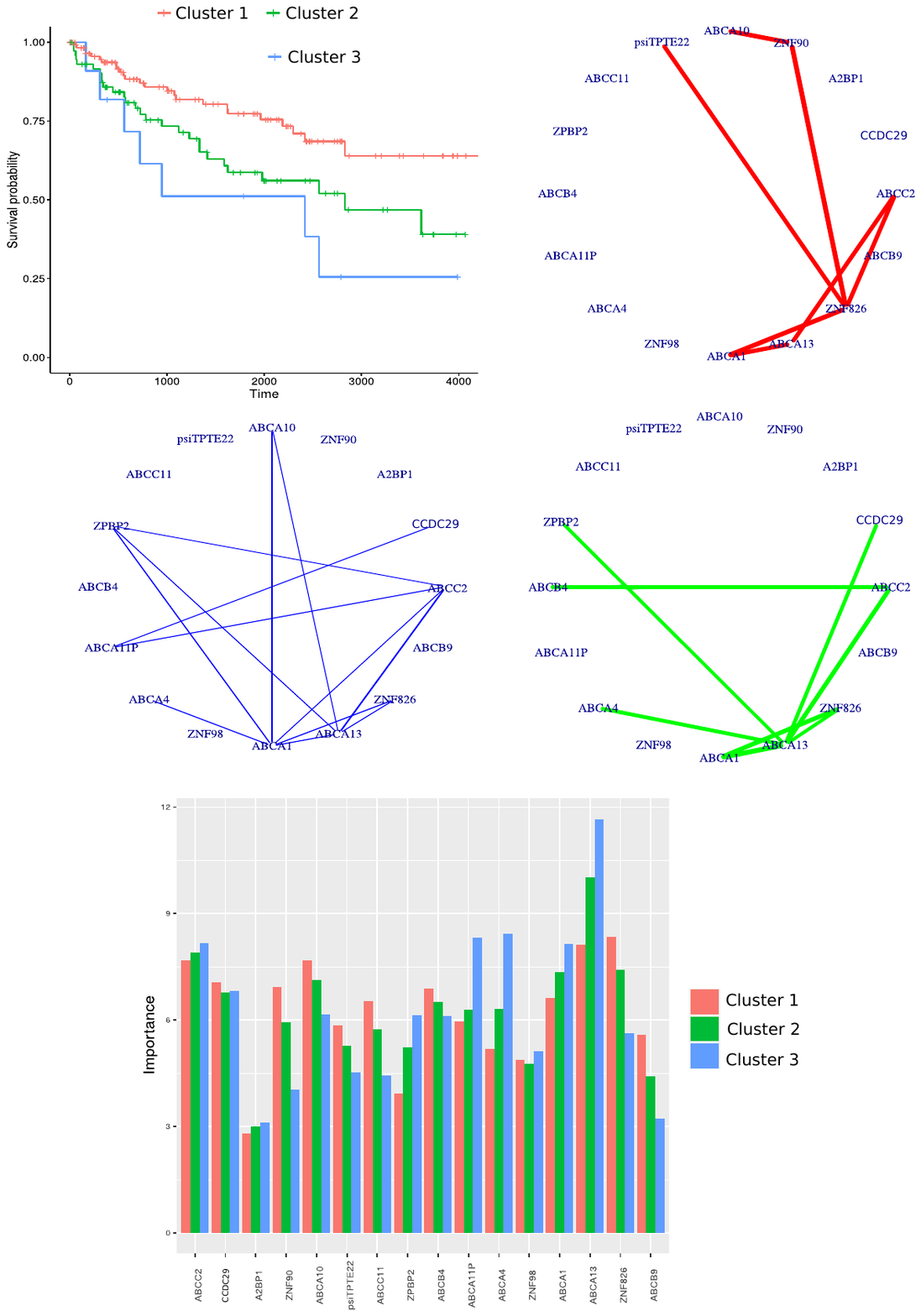}
    \caption{Application on TCGA data for interpretable disease subtype discovery.}
    \label{fig:TCGA_appl}
\end{figure}

The application on kidney cancer subtyping reveals three patient clusters, which we further analyse in terms of medical time-to-event patterns. The survival curves are significantly different among the three clusters with a log-rank $p$-value of $0.009$ (see \Cref{fig:TCGA_appl}). Compared to cluster 1, the risk for a death outcome is significantly higher for patients assigned to cluster 2 and cluster 3.  
Moreover, \Cref{fig:TCGA_appl} shows the cluster-specific feature graphs, where the edges with weights in the upper 0.95 quantiles are displayed. Interestingly, cluster 1 and cluster 2 consist of the same number of edges with an intersection of only three edges, namely ABCA1-ABCA13, ABCA1-ZNF826, and ABCA13-ABCC2. This suggests that the remaining interactions are unique to the detected disease subtypes. The feature importance of genes in the feature graphs are displayed at the bottom of \Cref{fig:TCGA_appl}. There is a notable difference in a gene called ABCA13, which is the most important gene in cluster 2 and cluster 3. In addition, a substantial difference between cluster 2 and cluster 3 can be observed in the genes ABCA11P and ABCA4, which are particularly important for cluster 3. 
The detected genes are part of the ATP Binding Cassette (ABC) transporters, extensively researched in cancer for their involvement in drug resistance. More recently, their impact on cancer cell biology has gained attention, with substantial evidence supporting their potential role across various stages of cancer development, encompassing susceptibility, initiation, progression, and metastasis~\cite{nobili2020role}.
However, further analysis is needed to draw meaningful medical conclusions. 


\section{Conclusion and future work}\label{sec:CF}

The study presents a novel method for constructing feature graphs from unsupervised random forests, applicable to both entire datasets and individual clusters. The constructed graphs exhibit desirable properties, with feature centrality reflecting their relevance to the clustering task and edge weights indicating the discriminative ability of feature pairs. Additionally, two feature selection strategies are presented: a brute-force approach with exponential complexity and a greedy method with polynomial complexity. Extensive evaluation on synthetic datasets demonstrates consistent behaviour between the greedy and brute-force methods, suggesting that, in the considered experimental setting, the greedy approximate algorithm also yields optimal solutions. 
Further evaluation on benchmark datasets reveals that the proposed greedy approach is more effective in selecting feature subsets than an impurity-based feature selection strategy, leading to superior clustering performance.
Notably, the greedy approach also exhibits a higher degree of monotonicity in performance, with clustering performance increasing with the number of selected features. This added property of the proposed approach can mitigate the negative impact of selecting a suboptimal number of features.
%
Moving forward, there are several avenues for further exploration. First, the proposed procedure for building feature graphs can be enhanced by defining alternative edge-building criteria, for instance, by combining sample and fixation strategies.
Furthermore, the constructed graph offers ample opportunities for exploration through a diverse range of graph metrics, including eccentricity, clustering coefficients, and various centrality metrics. Additionally, graph mining algorithms such as graph search and community detection methods could unveil deeper insights into the structure of the feature graph.
Finally, while the current graph-building procedure only considers splitting features, enriching the graph representation to also include splitting values could provide additional context to the feature interactions captured within the graph.

\bibliographystyle{splncs04}
\bibliography{bibliography}


\end{document}